\pdfoutput=1

\documentclass[11pt]{article}

\usepackage[final]{acl}

\usepackage{times}
\usepackage{times}
\usepackage{latexsym}

\usepackage[T1]{fontenc}

\usepackage[utf8]{inputenc}

\usepackage{microtype}

\usepackage{inconsolata}

\usepackage{graphicx}

\usepackage{graphicx}
\usepackage{amsmath}
\usepackage{multirow}
\usepackage{booktabs}
\usepackage{array}
\usepackage{subfigure}

\usepackage{amssymb,mathtools}
\usepackage{verbatim}
\usepackage{bm}     
\usepackage{enumitem}       
\usepackage{arydshln} 

\title{Lossless KV Cache Compression to 2\%}

\author{
    Zhen Yang\textsuperscript{\rm *,1},
    J. N. Han\textsuperscript{\rm *,1},
    Kan Wu\textsuperscript{\rm 1},
    Ruobing Xie\textsuperscript{\rm +,1}, \\
    \textbf{An Wang \textsuperscript{\rm 1,2},}
    \textbf{Xingwu Sun\textsuperscript{\rm 1},}
    \textbf{Zhanhui Kang\textsuperscript{\rm 1}}
    \\
    \\
    \textsuperscript{1} Tencent Hunyuan\\
    \textsuperscript{2} Tokyo Institute of Technology
}

\begin{document}
\maketitle

\def\thefootnote{*}\footnotetext{These authors contributed equally.}

\def\thefootnote{+}\footnotetext{Corresponding author}

\begin{abstract}

Large language models have revolutionized data processing in numerous domains, with their ability to handle extended context reasoning receiving notable recognition. To speed up inference, maintaining a key-value (KV) cache memory is essential. Nonetheless, the growing demands for KV cache memory create significant hurdles for efficient implementation. This work introduces a novel architecture, Cross-Layer Latent Attention (CLLA), aimed at compressing the KV cache to less than 2\% of its original size while maintaining comparable performance levels. CLLA integrates multiple aspects of KV cache compression, including attention head/dimension reduction, layer sharing, and quantization techniques, into a cohesive framework. Our extensive experiments demonstrate that CLLA achieves lossless performance on most tasks while utilizing minimal KV cache, marking a significant advancement in practical KV cache compression. 

\end{abstract}

\section{Introduction}

Large language models (LLMs) have been widely adopted and verified in various of fields, reshaping the way we collect and process information and impacting our daily lives \cite{driess2023palm,zhang2023recommendation,zhu2023large,wang2024survey}. Recently, the long-context reasoning and understanding abilities of LLMs have gradually been acknowledged to be essential in reflecting LLMs' capabilities and attracted more and more attention. Both closed-source and open-source LLMs are now striving to accommodate longer token lengths \cite{achiam2023gpt,deepseek2024v2}. However, this rapid expansion in length introduces critical efficiency challenges into LLMs, particularly concerning the growing \textbf{key-value (KV) cache} memory issue, which poses a significant barrier to the practical deployment of more powerful LLMs.
The KV cache technique, which involves caching and reusing previously computed key and value vectors from the classical multi-head attention (MHA) blocks \cite{vaswani2017attention} in decoder-only Transformers, is broadly adopted to accelerate model inference speed. This approach, however, comes at the cost of increased GPU memory usage and is considered a default configuration in most decoder-only LLMs \cite{bai2023qwen,touvron2023llama}.
Alleviating the increasing KV cache issue could free up GPU memory for other high-performing yet memory-intensive techniques, or alleviate constraints related to essential parameters such as batch size and token lengths, which are crucial for deploying more effective LLMs \cite{sheng2023flexgen,pope2023efficiently,ainslie2023gqa}.

With the rapid growth of LLM sizes, numerous studies have raised concerns regarding the expanding KV cache and proposed various solutions. The overall KV cache size is associated with several factors, including the number of attention heads, the number of layers, the dimension of each head, and the sequence length.
To enhance MHA efficiency in Transformers, methods such as multi-query attention (MQA)  \cite{shazeer2019fast} and grouped-query attention (GQA) \cite{ainslie2023gqa} are proposed to reduce KV cache requirements by decreasing the number of KV heads. Additionally, multi-head latent attention (MLA) has been developed to mitigate performance degradation through low-rank KV joint compression \cite{deepseek2024v2}. These methods concentrate more on the attention head and dimension aspects.
Conversely, another set of methods focuses on reducing the number of layers involved in KV cache. YOCO \cite{sun2024you} proposes a decoder-decoder architecture with linear attention that compresses the cache to a single layer. Similarly, CLA \cite{brandon2024reducing} and MLKV \cite{zuhri2024mlkv} share KV activation across adjacent layers, reusing KV cache of earlier layers.
Furthermore, there has been considerable interests in KV quantization techniques \cite{hooper2024kvquant,liu2024kivi}.

Despite these advancements, existing solutions often lead to non-negligible performance degradation in LLMs when subjected to significant KV cache compression ratios. Moreover, most research tends to concentrate on individual aspects of KV cache compression rather than effectively integrating multiple strategies into a cohesive architecture, which is efficient and non-trivial in practice.

In this work, we propose a novel \textbf{\underline{C}ross-\underline{L}ayer \underline{L}atent \underline{A}ttention (CLLA)} architecture for KV cache compression in LLM. Our objective is to reduce the KV cache to less than 2\% of its original size of classical MHA while preserving comparable performance levels. In CLLA, we explore the potential for integrating various aspects of KV cache compression methods into a unified and stable framework. Specifically, we jointly address KV cache compression through attention head/dimension aspect, layer aspect, and quantization aspect, and explore various possible combinations to achieve lossless KV cache compression.

Extensive results demonstrate remarkable effectiveness of our CLLA-quant: LLMs can achieve lossless performance on most tasks while utilizing less than 2\% of the original KV cache, with detailed analyses on our exploration on different CLLA variants. The primary contributions of this work are summarized as follows:
\begin{itemize}[leftmargin=*]
    \item We introduce CLLA and CLLA-quant, a novel architecture for KV cache compression that intelligently integrates head/dimension, layer, and quantization aspects.
    \item We explore diverse variants of CLLA for better performance. To our knowledge, CLLA is the first architecture to achieve a KV cache compression ratio below 2\% with lossless performance.
    \item We conduct extensive in-depth explorations and analyses on various ablations and possible combinations of KV cache compression methods, which could shed lights on future research and practical applications. We hope our simple and effective CLLA could facilitate LLM community.
\end{itemize}

\section{Related Works}

\subsection{Large Language Models}

Recently, transformer-based models have garnered significant attention due to their powerful performance in various fields, including NLP and multimodal tasks. Large language models such as GPT-4 \cite{openai2023gpt} and LLaMA3 \cite{llama3modelcard} have achieved remarkable results across a range of natural language processing tasks.

However, the traditional decoder-only architecture incurs substantial training and online inference costs. To address this issue, Palm \cite{chowdhery2023palm} and StarCode \cite{li2023starcoder} utilize MQA \cite{shazeer2019fast}, where all attention heads share a group of KV. This approach aims to reduce the computational overhead associated with the attention mechanism. LLaMA2 \cite{touvron2023llama} employs a different strategy within its attention blocks by grouping keys and values (GQA) \cite{ainslie2023gqa} to minimize the KV cache during inference. Concurrently, it increases the feed-forward network hidden size to maintain model performance. FlashAttention \cite{dao2023flashattention} optimizes training and inference efficiency from the perspective of GPU I/O operations. This method specifically targets the bottlenecks in data movement and computation within the GPU architecture. Mixture-of-Experts models~\citep{jiang2024mixtral,muennighoff2024olmoe,wang2024hmoe} utilize sparse architectures to achieve improved performance.

\subsection{KV Cache Compression}

Previous research has utilized techniques such as reducing the dimensions of attention heads and employing low-rank matrix approximations to enhance model efficiency. Some recent work reduces the KV cache through the layer dimension\citep{wu2024layer,zuhri2024mlkv,he2024efficient,liu2405minicache}. Notably, Yu effectively compressed models by minimizing errors in MHA-to-GQA transitions \cite{yu2024effectively}, while maintaining compatibility with RoPE \cite{su2023enhanced}. Some research efforts concentrate on reducing KV cache requirements by decreasing the sequence length, using methods such as token dropping \cite{zhang2024h2o,li2024snapkv,xiao2023efficient,shi2024discovering} and prompt compression \cite{jiang2023llmlingua,chuang2024learning}. However, these methods often lead to a notable decline in performance.

Recently, many research efforts have employed quantization compression to reduce KV cache, thereby improving inference efficiency \citep{banner2018scalable,wu2020easyquant,yang2024no,he2024zipcache,liu2024unlocking}. Gear \cite{kang2024gear} utilizes low-rank matrices and low-precision quantization to achieve near-lossless 4-bit KV cache compression. Coupled Quantization \cite{zhang2024kv} encodes multiple KV channels into a low-bit code. KIVI \cite{liu2024kivi} performs quantization on keys and values along the channel and token dimensions, respectively. These works have primarily focused on inference.
In contrast, we propose CLLA-quant method to quantize the KV cache in 4-bit integers during training and deployment.

\begin{figure}[t]
\centering
\includegraphics[width=0.5\textwidth]{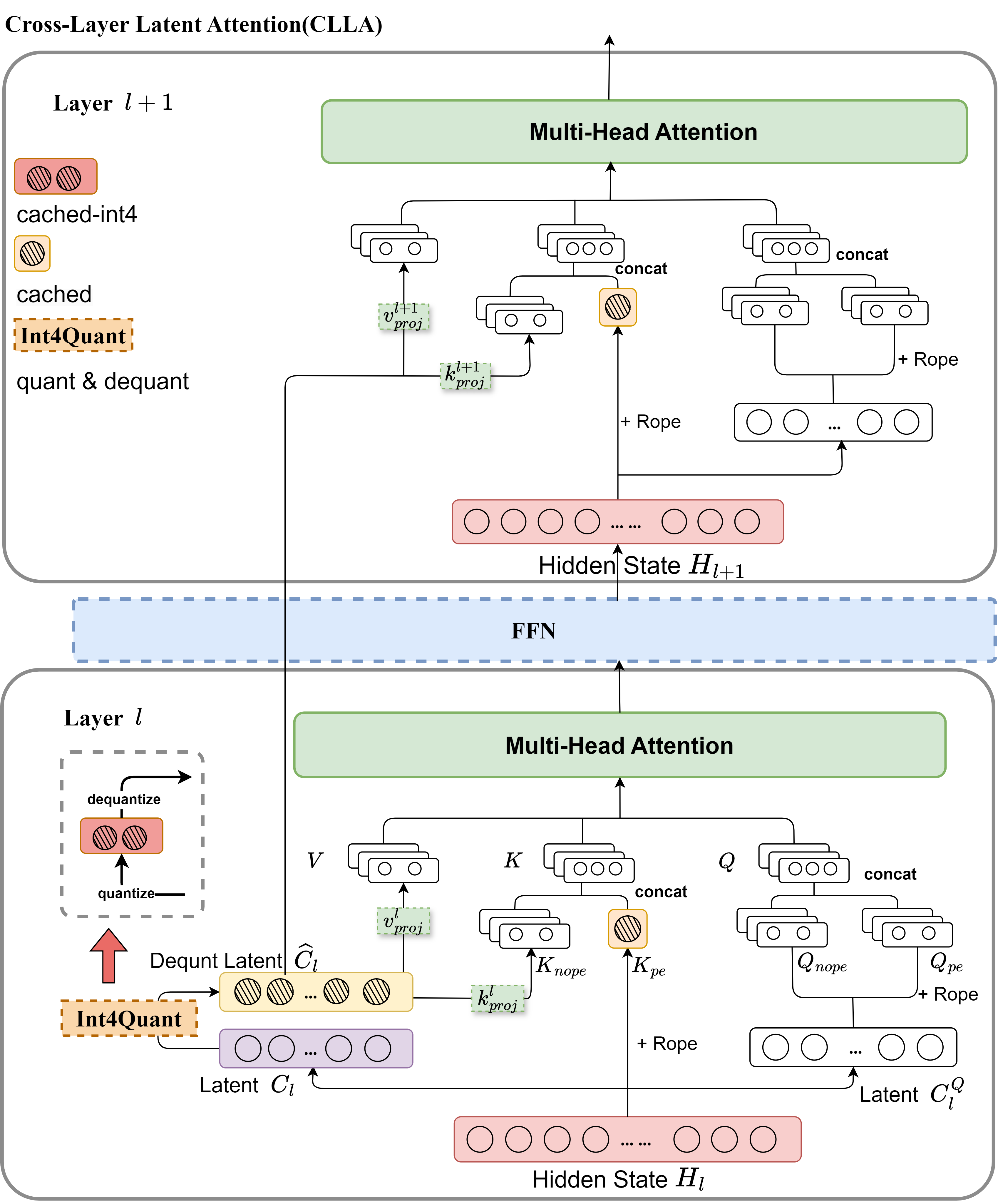}
\caption{Overview of the Cross-Layer Latent Attention (CLLA) architecture. During training with CLLA, the preceding layer computes the compressed key-value (KV) pairs after quantization and dequantization. The subsequent layer then continues to utilize the compressed KV from the previous layer. During inference, the model stores the compressed KV pairs of the l-th layer with int4 quantization.}
\label{CLLA_overview.png}
\end{figure}

\section{Methodology}

\subsection{Preliminary of KV Cache}

Currently, transformer model is the most prevalent LLM architecture. It is composed of many blocks stacked together. Each block includes an attention layer and a MLP layer. Multi-head Attention (MHA) \cite{vaswani2017attention} is a typical design for the attention layer. Given the sequence hidden states $\mathbf{H} = [h_1, h_2, ..., h_t]$, the computations of MHA for each head are performed as follows:
\begin{equation}
\begin{aligned}
&\text{MHA}(\mathbf{H}) = \text{Concat}(\text{head}_1, ..., \text{head}_n)\mathbf{W}^O, \\
&\text{head}_i = \text{Att}(\mathbf{Q}_i, \mathbf{K}_i, \mathbf{V}_i), \quad i = 1, \ldots, N, \\
&\mathbf{Q}_i = \mathbf{H}\mathbf{W}_i^Q, 
\mathbf{K}_i = \mathbf{H}\mathbf{W}_i^K,
\mathbf{V}_i = \mathbf{H}\mathbf{W}_i^V, \\
\end{aligned}
\end{equation}
where $\mathbf{W}_i^Q$, $\mathbf{W}_i^K$, and $\mathbf{W}_i^V$ are weight matrices in $\mathbb{R}^{D_h \times D_h}$ for $\text{head}_i$. $N$ denotes the number of heads. $\text{Att}(\cdot)$ represents the calculation of scaled dot-product attention \cite{vaswani2017attention}. 

To increase the speed of autoregressive decoding, it is necessary to store the key and value vectors ($K$ and $V$) of previous input hidden states. In Transformer, the memory requirement for KV cache is influenced by multiple parameters: batch size ($B$), number of layers ($L$), sequence length ($S$), number of attention heads ($N$), and the dimension of each head ($D_h$). This dependency can be formulated as:
\begin{equation}
        \text{Memory}(\text{KV Cache}) = \mathcal{O}(2BLSND_h).
\end{equation}

Classical methods such as GQA \cite{ainslie2023gqa}, MLA \cite{deepseek2024v2}, and CLA \cite{brandon2024reducing} usually compress KV cache from $N$, $D_h$, and $L$ aspects as shown in Table \ref{tab:Cache Memory}. In this work, we strive to jointly consider all these aspects for performance-lossless KV cache compression.

\begin{table}[t]
        \centering
                \begin{tabular}{lc}
                        \toprule
                        Method & KV Cache Memory \\
                        \midrule
                        MHA & $\mathcal{O}\left(BSLND_h\right)$ \\
                        GQA & $\mathcal{O}\left(BSLGD_h\right)$ \\
                        MLA & $\mathcal{O}\left(BSLC\right)$ \\
                        CLA & $ \mathcal{O}\left(BSLND_h / F\right)$ \\
                        CLLA & $\mathcal{O}\left(BSLC / F\right)$ \\
                        CLLA-quant & $\mathcal{O}\left(BSLC / 4F\right)$ \\
                        \bottomrule
                \end{tabular}
        \caption{Theoretical KV cache memory complexity for different methods for a sample. $B$, $S$, $L$, $N$, $G$, $D_h$, are the batch size,  sequence length, layer number, KV head number, GQA group size, and head dimension. $C$ and $F$ denote MLA KV latent hidden size and CLA sharing factor, respectively.}
        \label{tab:Cache Memory}
\end{table}

\begin{figure*}[!htpb]
\centering
\includegraphics[width=1\textwidth]{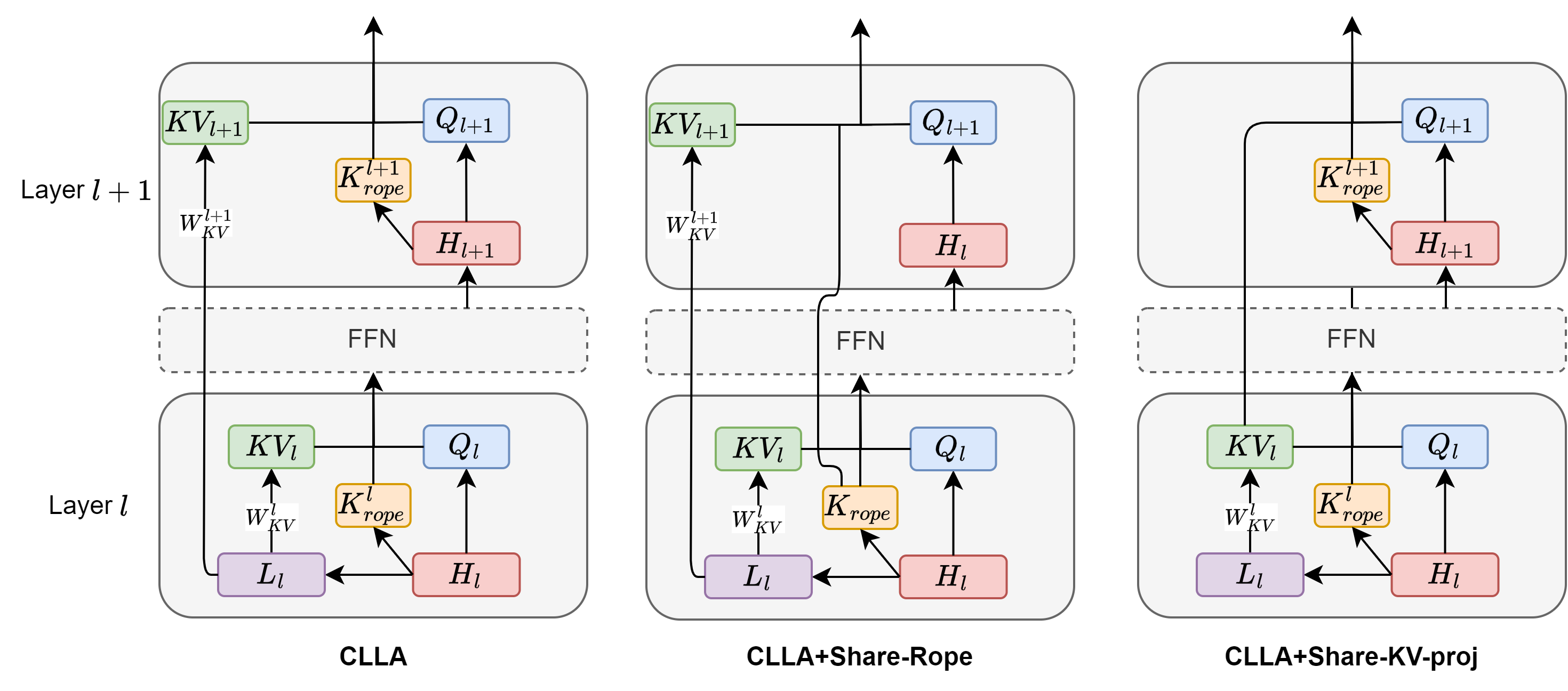}
\caption{Three different approaches for passing latent key-value (KV) pairs and K-rope in the attention mechanism to the next layer within the CLLA architecture. The final CLLA selects the left version (CLLA+share-latent).}
\label{clla_ablation.png}
\end{figure*}

\subsection{CLLA: CLA Meets MLA}

CLLA involves compressing KV cache into low-rank latent states and subsequently allowing multiple layers to share these states. The key research point is how to effectively share compressed latent states across layers.

\noindent
\textbf{MLA basics.}
To begin with, we project the input hidden states of the attention layer into low-rank latent states denoted as $\mathbf{C}$:
\begin{equation}
\begin{aligned}
&\mathbf{C} = \mathbf{H}\mathbf{W}^C. \\
\end{aligned}
\end{equation}
This method enables significant compression since only the latent states 
$C$ need to be cached, rather than storing entire key and value vectors. During attention computation, we restore the latent states back to the standard key-value representations.
\begin{equation}
\begin{aligned}
&\mathbf{K} = \mathbf{C}\mathbf{W}^K, \quad
&\mathbf{V} = \mathbf{C}\mathbf{W}^V. \\
\end{aligned}
\end{equation}
Subsequently, we compute the query vector to engage in attention calculation. To facilitate the use of RoPE, two additional small query vector $Q_{rope}$ and key vector $K_{rope}$ are created as follows,
\begin{equation}
\begin{aligned}
    \mathbf{Q} &= \mathbf{H}\mathbf{W}^{Q}, \\
    \mathbf{Q}_{\text{rope}} &= \mathbf{H}\mathbf{W}^{Q_{\text{rope}}}, \\
    \mathbf{K}_{\text{rope}} &= \mathbf{H}\mathbf{W}^{K_{\text{rope}}}.
\end{aligned}
\end{equation}
Lastly, we incorporate vectors containing RoPE information into the query and value vectors and proceed with the attention calculation.
\begin{equation}
\begin{aligned}
&\mathbf{Q}' = \text{Concat}(\mathbf{Q}, \mathbf{Q}_{\text{rope}}), \\
&\mathbf{K}' = \text{Concat}(\mathbf{K}, \mathbf{K}_{\text{rope}}), \\
&\textbf{MLA}(\mathbf{C},\mathbf{V}) = \text{MH}(\text{Att}(\mathbf{Q}', \mathbf{K}', \mathbf{V})),
\end{aligned}
\end{equation}
where $\text{MH}(\cdot)$ denotes the operation that implements the multi-head mechanism.

\noindent
\textbf{CLLA variants.}
When integrating MLA and CLA strategies, several variations and complexities must be considered. Compared to the naive CLA method (which shares KV activation directly across layers), we depict three variations in Figure \ref{clla_ablation.png}. The left variation is adopted as our final CLLA architecture. It shares only latent vectors across layers while allowing each layer its own KV projection matrices to reconstruct KV activation. This variation demonstrated superior performance compared to others, and detailed experiments are presented in Section \ref{sec: sharing strategy}.
Formally, for CLLA with a sharing factor of 2, the equation is defined as follows,
\begin{equation}
O_l = 
\begin{cases}
    \text{MLA}(\textbf{C}_l, \textbf{V}_l), & \text{if }  l\bmod2 = 0 \\
    \text{MLA}(\textbf{C}_{l-1}, \textbf{V}_l), & \text{if }  l\bmod2 = 1
    .
\end{cases}
\end{equation}
In this formulation, we share latent vectors while each layer utilizes its own RoPE vectors for keys. For layers employing latent vectors from previous layers, specific linear layers are used to project shared latent vectors back into $K_{latent}$ and $V$. Subsequently, $K_{latent}$ is concatenated with $K_{rope}$ for MHA calculations alongside $Q$ and $V$.

\noindent
\textbf{Explanation of the current CLLA selection.}
Although this approach requires more computational resources compared to variations that share both KV activation but not latent vector $C$, we believe this trade-off is justified by its superior performance relative to other methods. We hypothesize that this improvement arises because each projection matrix can tailor latent vector projections into specific KV activation containing relevant information for that layer. Furthermore, we found that sharing only latent vectors significantly enhances performance when combined with quantization techniques due to quantization being applied at the latent vector level. Similar reasoning applies regarding specific projection layers mitigating quantization loss.

\subsection{CLLA-quant: Latent Quantization}

\DeclarePairedDelimiter{\nint}\lfloor\rceil
\DeclarePairedDelimiter{\abs}\lvert\rvert
To further reduce memory overhead associated with KV cache, we implement low-bit quantization on latent states by converting them from 16-bit floating-point values into lower-bit integers using symmetric quantization \cite{nagel2021white}. The quantization function can be expressed as follows:
\begin{equation}
    Q\left(\mathbf{C}\right) = \text{int}_{b}\left(\text{clip}\left( \nint*{\frac{\mathbf{C}}{S}}, I_{min}, I_{max}\right)\right),
\end{equation}
\begin{equation}
S = \text{max}(\abs{\mathbf{C}}) / I_{max},
\end{equation}
\begin{equation}
    \text{clip}\left(x, a, b\right) = max\left(a, min\left(b, x\right)\right).
\end{equation}
Here $\mathbf{C}$ represents the latent states, and $Q\left(\mathbf{C}\right)$ represents their quantized version stored as $b$-bit integers. $\nint{\cdot}$ denotes the rounding operator, while $S$ serves as the scaling factor. The latent states are scaled to the restricted range $\left[I_{min}, I_{max}\right]$, which is set to $\left[-\left(2^{b-1} - 1\right), 2^b - 1\right]$. The quantized latent states are stored in cache and dequantized only when needed during computations. The dequantization function is defined as follows,
\begin{equation}
    \hat{\mathbf{C}} = S\times \text{float}\left(Q\left(\mathbf{C}\right)\right).
\end{equation}
The recovered latent states $\mathbf{\hat{C}}$ are then utilized in computations. We adopt 4-bit for CLLA-quant.

\noindent
\textbf{Quantization recipe.}
To minimize quantization error further, we employ sub-channelwise quantization techniques \cite{survey_quant}, where every $g$ elements are grouped, with each group sharing a scaling factor. Additionally, prior to quantization, latent states are normalized using RMSNorm \cite{rmsnorm}.
Both quantization and dequantization processes are applied during training and deployment phases. In training mode specifically, gradients associated with quantization are approximated using the straight-through estimator (STE) \cite{ste}.

\section{Experiments}
\subsection{Datasets and Experimental Settings}

The training dataset was collected in-house and consists of a cleaned combination of Chinese and English datasets. We trained all models on 300 billion tokens. For evaluation, we aim to achieve robust conclusions across diverse domains in both English and Chinese. We conducted comprehensive evaluations involving 7 English tasks, including MMLU, Hellaswag, ARC-C, Winogrande, BoolQ, PIQA, and SIQA. For Chinese tasks, we evaluated models on 8 tasks: CMMLU, CEval, CLUEWSC2020, Thuc-news, OCNLI, C3, FewCLUE-CHID, and CMNLI. More details are noted in Appendix \ref{app:datasets detailss}.

In order to ensure equitable comparison, the intermediate dimensionality of the feed-forward neural network (FFN) was adjusted to maintain comparable activation parameters across all models. The Appendix elaborates on the implementation specifics of each model. The training process was executed utilizing 64 NVIDIA H800 GPUs, with the same hardware resources employed for the inference evaluations. Complete details of the training configurations are provided in the Appendix.

\begin{table*}[t]
        \centering
        \resizebox{0.90\linewidth}{!}{
                \begin{tabular}{lccccccccc}
                        \toprule
                        Model & MMLU & Hellaswag & ARC-C & Winogrande & BoolQ  & PIQA & SIQA & AVG\\
                        \midrule
                        MHA & 38.61 & 54.78  & 48.10  & 55.31 & 70.60  & 73.76 &  55.94 &  56.73 \\
                        GQA & 37.59 & 54.10 & 51.51  & \textbf{56.91} & 69.67  & \textbf{74.70} & 53.33 & 56.83 \\
                        MLA & 40.51 & 54.23 & 48.83 & 55.56 & 70.83  & 73.94 & 56.19 & 57.16 \\
                        \midrule
                        CLLA & \textbf{42.08} & \textbf{54.60} & \textbf{52.84} & 54.38 & \textbf{73.10} & 74.43 & 54.35 & 57.97 \\
                        CLLA-quant & 41.13 & 53.87 & 51.84 & 55.64 & 72.33 & 74.32 & \textbf{56.70} & \textbf{57.98} \\
                        \bottomrule
                \end{tabular}
        }
        \caption{Main results on English benchmarks.}
        \label{tab:eng benchmarks}
\end{table*}

\begin{table*}[t]
        \centering
        \resizebox{\linewidth}{!}{
                \begin{tabular}{lcccccccccc}
                        \toprule
                        Model & CMMLU & CEval & CLUEWSC2020 & Thuc-news & OCNLI & C3 & CHID & CMNLI & AVG\\
                        \midrule
                        MHA & 36.88 & 42.78 & 69.42 & 96.14 & 55.61 & 60.43  & 79.20  & 52.79 & 61.66 \\
                        GQA & 36.79 & 39.99 & 70.19 & 95.47 & \textbf{56.14} & 60.40 & 78.71 & \textbf{54.55} & 61.53 \\
                        MLA & 39.20 & 44.29 & 69.23 & 96.27 & 54.07 & 61.20 & 78.71 & 50.78 & 61.72 \\
                        \midrule
                        CLLA & 40.87 & \textbf{44.92} & \textbf{73.08} & 96.10 & 55.39 & \textbf{61.77} & \textbf{82.67} & 53.67 & \textbf{63.56}  \\
                        CLLA-quant & \textbf{44.82} & 44.63 & 70.19 & \textbf{96.47} & \textbf{56.14} & 59.87 & 79.70 & 54.27 & 63.26 \\
                        \bottomrule
                \end{tabular}
        }
        \caption{Main results on Chinese benchmarks.}
        \label{tab:zh benchmarks}
\end{table*}

\subsection{Competitors}
\label{experiments}

For every competitor, we utilize the same 1.44 billion activation parameters with a Mixture of Experts (MoE) architecture \cite{fedus2022switch,lepikhin2020gshard}. Specifically, we employ a model with a 1Bx16 MoE configuration, comprising nearly 11 billion parameters in total, to ensure fair comparisons.The following methods were evaluated:

\textbf{MHA}: The foundational transformer architecture \cite{vaswani2017attention,radford2019language} resembles that of the Llama2 model \cite{touvron2023llama}. It features SwiGLU \cite{shazeer2020glu} as the activation function for the feed-forward layer and utilizes RoPE \cite{su2023enhanced} for the embedding of relative positions. All subsequent models are variations of this foundational model.

\textbf{GQA}: GQA \cite{ainslie2023gqa} groups query heads together, allowing each group to share keys and values. We set the number of KV heads to 8 out of a total of 16 query heads to reduce KV cache by half.


\textbf{MLA}: In MLA \cite{deepseek2024v2}, the KV cache is mapped into a low-rank space. The latent dimension for this low-rank space is set at 512.

\textbf{CLLA}: Our method combines the settings from MLA and CLA \cite{brandon2024reducing} while adhering to the aforementioned sharing configurations.

\textbf{CLLA-quant}: We applied int4 scalar quantization to compress latent states with a quantization group size of 32 on the original CLLA.

In each model, the hidden size is configured to 1536, with 32 layers and 16 attention heads. We incorporated 17 experts in the MoE layer, consisting of one shared expert alongside 16 experts, from which the top-1 is selected for routing. To ensure the alignment of activation parameters and total parameters across various methods, we modified the expansion ratio of the feed-forward network (FFN) as needed for fair comparisons. Further details regarding training and model hyperparameters can be found in Appendix \ref{app:implementation detailss}.

\subsection{Main Results}

After pre-training on 300 billion tokens, we evaluated all models across 15 benchmarks that encompass a wide range of tasks.
All models are compared in the same tokenizer and data. 
On English benchmarks shown in Table \ref{tab:eng benchmarks}, MLA are comparable with GQA baseline. The proposed models CLLA and CLLA-quant achieve better results with higher compress ratio. We can achieve similar conclusion on Chinese benchmarks presented in Table \ref{tab:zh benchmarks}.
Additionally, theoretical KV cache memory bytes usages for each model are shown in Table \ref{tab:detail bytes}.
From these tables, several interesting observations emerge regarding the effectiveness of our methods in achieving high KV cache compression ratios without performance degradation:

\begin{table}[t]
        \centering
        \small
                \begin{tabular}{lcc}
                        \toprule
                        Method & KV Cache Memory & Compression Ratio \\
                        \midrule
                        MHA & 6,144 Bytes & 100\% \\
                        GQA & 3,072 Bytes & 50.0\% \\
                        MLA & 576 Bytes & 9.4\%\\
                        CLLA & 320 Bytes & 5.2\% \\
                        CLLA-quant & 128 Bytes & 2.1\% \\
                        \bottomrule
                \end{tabular}
        \caption{The actual cache memory costed for each tokens of compared 1B models and the compression ratio. Conducting the inference with 16-bit precision implies that each scalar within the KV cache occupies 2 bytes.
        }
        \label{tab:detail bytes}
\end{table}

\begin{table*}[!htbp]
        \centering
        \resizebox{\linewidth}{!}{
                \begin{tabular}{lccccccc}
                        \toprule
                        Sharing Method & Model & $d_{head}$ & Sharing Factor & Q Heads & KV Heads & KV Bytes/Token $\downarrow$ & Val PPL $\downarrow$ \\
                        \midrule

                        \multirow{4}*{Intra-layer} & GQA-Group8 & 96 & 1 & 16 & 8 & 49,152 & 8.93 \\
                         ~ & GQA-Group4 & 96 & 1 & 16 & 4 & 24,576 & 8.95 \\
                         ~ & GQA-Group2 & 96 & 1 & 16 & 2 & 12,288 & 9.05 \\
                         ~ & MQA & 96 & 1 & 16 & 1 & 6,144 & 9.11 \\
                         \midrule
                         
                         \multirow{3}*{Inter-layer} & CLA-Share2 & 96 & 2 & 16 & 16 & 49,152 & 8.97 \\
                        ~ & CLA-Share4 & 96 & 4 & 16 & 16 & 24,576 & 9.07 \\
                        ~ & CLA-Share8 & 96 & 8 & 16 & 16 & 12,288 & 9.47 \\
                        
                        \midrule
                         \multirow{4}*{Mixed} & GQA-Group4+CLA-Share2 & 96 & 2 & 16 & 4 & 12,288 & 9.12 \\
                         ~ & GQA-Group2+CLA-Share2 & 96 & 2 & 16 & 2 & 6,144 & 9.19 \\
                         ~ & MQA-Dim384-CLA-Share2 & 384 & 2 & 4 & 1 & 12,288 & 9.12 \\ 
                         ~ & MQA-Dim192-CLA-Share2 & 192 & 2 & 8 & 1 & 6,144 & 9.16 \\
                        \bottomrule
               \end{tabular}
        }
        \caption{Ablation results for intra-layer and inter-layer sharing strategies with the same model size. 
        }
        \label{tab:cla ablation}
\end{table*}

First, CLLA notably surpassed MHA in performance while also achieving a significant reduction in memory usage of nearly 95\%. For English tasks, CLLA's performance was about 1.24 point higher than that of MHA, and it showed similar enhancements for Chinese tasks. This improvement is possibly attributed to our model design. Under the same amount of activated parameters,
we were able to increase the FFN hidden size by utilizing parameters saved from the CLLA approach, which likely contributed to the improved outcomes.

Second, CLLA-quant exhibited results comparable to CLLA while further achieving a fourfold reduction in memory usage. These results verify the impressive conclusion that our CLLA-quant method compresses KV cache memory to 2\% without performance loss generally on 15 benchmarks compared to MHA. We hypothesize that the minimal impact of quantization on performance is due to its application during pretraining, allowing the entire model to adapt to the quantized KV cache. The quantization likely enforces a form of robustness in the representations stored in the KV cache, ensuring that minor precision losses do not adversely affect the model's ability.

\section{In-depth Model Analyses on CLLA} \label{ablation}

It is nontrivial to compress KV cache from different aspects including layer, dimension, and bit levels. In this section, we conduct extensive explorations on different (successful or failed) strategies and report the insights found in our experiments, hoping to facilitate future research.

\subsection{Intra-layer \emph{vs.} Inter-layer Sharing} 
To gain a deeper understanding of the strategy for combining CLA and MLA, we conduct a thorough analysis of both intra-layer and inter-layer sharing methods.
The superiority between inter-layer sharing methods, such as CLA, and intra-layer sharing strategies, such as MQA and GQA, has been a subject of contradictory conclusions in current studies. In \citet{brandon2024reducing}, the authors claim that with proper design, the combination of intra-layer and inter-layer strategies can make inter-layer sharing better than intra-layer sharing in some situations. However, a later study \cite{zuhri2024mlkv} found that intra-layer strategies are generally better than inter-layer strategies.
To explore these design choices, a series of experiments were conducted. The experimental models are in the same setting with our main experiment, except we only train 50B tokens for a fast comparison on validation perplexity. The results are presented in Table \ref{tab:cla ablation}:

(1) Intra-layer strategies consistently achieve lower validation perplexity compared to inter-layer strategies when using the same memory budget. Specifically, comparing GQA-Group8 with CLA-Share2, GQA-Group4 with CLA-Share4, and GQA-Group2 with CLA-Share8, intra-layer strategies consistently outperform inter-layer strategies in terms of validation perplexity. Furthermore, even with equivalent memory reductions, the performance loss associated with the CLA strategy is significantly greater than that of the GQA strategy. This conclusion is supported by comparisons between CLA configurations with varying share factors and GQA configurations with different group sizes.

(2) Neither the combination of GQA and CLA strategies nor adjustments to the head dimensions improved the performance. Specifically, when we combined both strategies, we observed that GQA-Group2+CLA-Share2 performed worse than MQA, while GQA-Group4+CLA-Share2 was inferior to GQA-Group2. Additionally, maintaining the total hidden dimension while adjusting head dimensions failed to surpass simple intra-layer sharing strategies. This is evident from comparisons between MQA-Dim192-CLA-Share2 and MQA as well as MQA-Dim384-CLA-Share2 and GQA-Group2.

Based on these observations, we can outline a design strategy for utilizing intra-layer and inter-layer sharing to conserve KV-cache memory. First, prioritize intra-layer strategies if the KV-cache memory budget can be met solely through intra-layer approaches. Second, resort to inter-layer strategies only after significantly compressing intra-layer memory strategies, such as MQA and MLA, when further reduction is necessary.

\begin{figure}[t]
\centering
\resizebox{\linewidth}{!}{
\includegraphics[width=0.98\columnwidth]{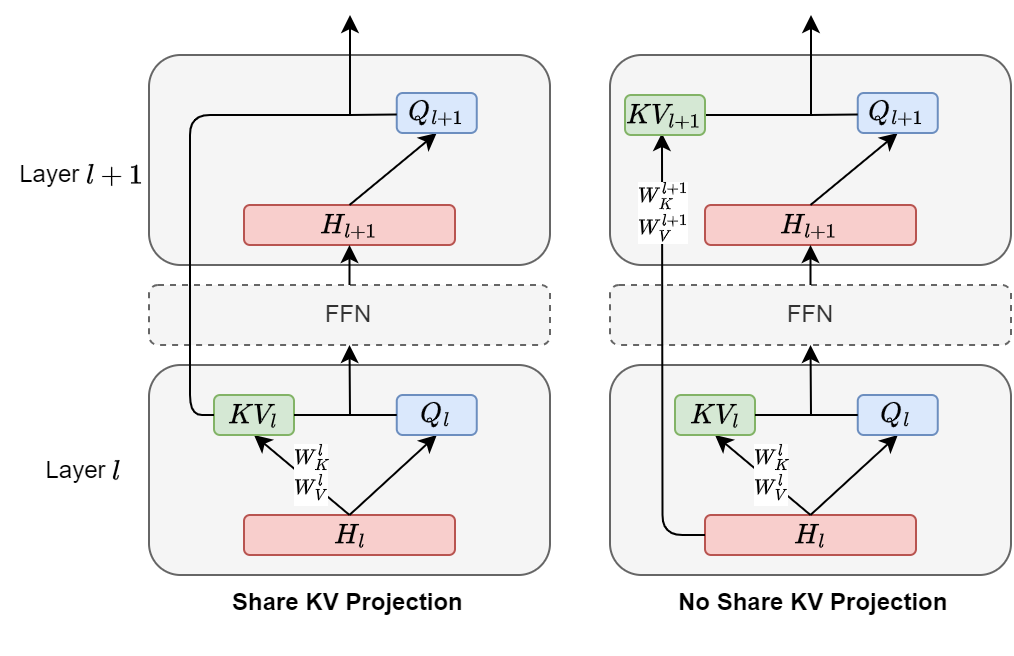}
}

\caption{\textbf{Left}: the model with KV projection weights sharing. \textbf{Right}: the model without KV projection weights sharing.}
\label{cla-proj.png}
\end{figure}

In addition to the aforementioned ablation studies, inspired by MLA, we explored the possibility of sharing input hidden states for attention blocks instead of KV heads. This approach necessitates caching input hidden states for inference while computing KV heads in real-time. We can view the MLA strategy as a LoRA variant of this concept. When applying CLA within this framework, we must determine whether to share KV projection weights as in Figure \ref{cla-proj.png}. Our findings indicate that allowing each layer its own KV projection weights yields slightly better performance than sharing KV projections, with validation perplexities of 8.97 versus 8.99. Although this decision is critical for CLLA, the difference is not substantial in this context.

\subsection{CLLA Cross-layer Sharing Strategies} \label{sec: sharing strategy}

Upon decomposing the components eligible for cross-layer sharing within the MLA strategy, we identified three elements: KV latent vectors, K rope vectors, and KV projection matrices. These three configurations are illustrated in Figure \ref{clla_ablation.png}. Our final design decision favors the first configuration due to its superior performance. The second configuration offers additional KV-cache savings but only marginally so because of the relatively small size of K rope vectors compared to KV latent vector sizes. The third configuration may reduce computation and model parameters.

We evaluated these three models based on downstream tasks under identical experimental settings as those used in the main experiments. The results are presented in Table \ref{tab:clla_ablation}, which displays average scores for both English and Chinese tasks. Detailed results for all English and Chinese task can be found in the Appendix \ref{app:ablation exp detailss}.

\begin{table}[t]
        \centering
        \resizebox{\linewidth}{!}{
                \begin{tabular}{lccc}
                        \toprule
                        Model & English & Chinese & AVG \\
                        \midrule
                        CLLA & 57.97 & \textbf{63.56} & \textbf{60.77} \\
                        CLLA+share-krope & 56.24 & 61.71 & 58.98 \\
                        CLLA+share-kvproj & \textbf{59.47} & 61.40 & 60.44 \\
                        \bottomrule
                \end{tabular}
        }
        \caption{The downstream results of CLLA sharing ablation experiments.}
        \label{tab:clla_ablation}
\end{table}

Our analysis reveals that CLLA with exclusive sharing of KV latent vectors achieved the best performance. The strategy of sharing K rope vectors significantly detracted from performance with minimal benefits. While the KV projection sharing strategy yielded better results for English tasks, it was lower overall. Nevertheless, this approach can provide moderate improvements during training and inference by eliminating half of the KV projection calculations and matrices required. Given our primary focus on downstream performance, we opted for the first configuration as our CLLA strategy.
Another reason for selecting this configuration is its favorable interaction with quantization techniques.

\subsection{CLLA Quantization Strategies}

Based on our previous findings, we selected CLLA and CLLA+share-kvproj for latent quantization training and compared their downstream performance as shown in Table \ref{tab:quant_ablation}. All experimental settings were consistent with those used in the main experiments.
Our results indicate that the strategy without sharing KV projection matrices achieved significantly higher performance when combined with quantization techniques.
The observed differences may be attributed to the fact that having multiple KV projections can better compensate for quantization loss compared to relying on a shared KV projection.

\begin{table}[t]
    \centering
    \resizebox{\linewidth}{!}{
                \begin{tabular}{lccc}
                        \toprule
                        Model & English & Chinese & AVG \\
                        \midrule
                        CLLA-quant & 57.98 & \textbf{63.26} & \textbf{60.62}\\
                        CLLA-quant+share-kvproj & \textbf{58.44} & 61.45 & 59.95 \\
                        \bottomrule
                \end{tabular}
    }
        \caption{The downstream results of different CLLA sharing strategy with quantization training.}
        \label{tab:quant_ablation}
\end{table}

\section{Conclusion and Future Work}

In this work, we introduce CLLA and CLLA-quant, two effective techniques for compressing KV caches. Both methods achieve substantial reductions in KV cache size without compromising performance, with CLLA-quant achieving memory usage as low as 2\% while maintaining performance comparable to the baseline. We identify optimal design choices that effectively combine attention head/dimension reduction, layer sharing, and quantization techniques, which is verified by extensive experiments and analyses. Although CLLA is straightforward, it successfully confirms the feasibility of lossless large-ratio KV cache compression, which is promising to be the superior fundamental component for efficient LLM inference.

In the future, we will explore our CLLA series on larger scales of LLMs to verify its universality. Further experiments will also be conducted to discover more efficient parameter allocations on different components of LLM's KV cache.

\section*{Limitations}
Limitations of the current work are discussed below.
Firstly, while the proposed CLLA-quant performs well in terms of achieving extreme KV cache savings, its structure is relatively complex. Streamlining this process while maintaining its effectiveness would greatly facilitate the practical application of this technology. Secondly, the experiments in this paper mainly focus on compressing and saving the KV cache. Further optimization of the framework is needed to reduce the time complexity in the Attention Block, ensuring high generation efficiency and throughput as the sequence length increases.

\bibliography{reference}
\clearpage

\appendix
\section{Implementation Details}
\label{app:implementation detailss}
To ensure a fair comparison, we primarily adjusted the intermediate size of the feed-forward neural network (FFN) to maintain similar activation parameters across models. The implementation details for each model are presented in Table \ref{tab:model_detail}. Aside from the parameters listed here, all other parameters are set to their default values. For instance, all models were configured with 32 layers and a vocabulary size of 128,256. In the MLA strategy, we established the KV LoRA rank at 512, the Q LoRA rank at 768, and the query-key rope head dimension at 64.

\begin{table*}[t]
        \centering
        \resizebox{\linewidth}{!}{
                \begin{tabular}{lcccccccc}
                        \toprule
                        Model & Intermediate Size & q heads & kv heads & head dim & topk/experts & Activation Parameters (B) & Total Parameters (B)\\
                        \midrule
                        MHA & 3840 & 16 & 16 & 96 & 1+1 / 1+16 & 1.44 & 10.13 \\
                        GQA & 4096 & 16 & 8 & 96 & 1+1 / 1+16 & 1.44 & 10.69 \\
                        MLA & 3840 & 16 & 16 & 96 & 1+1 / 1+16 & 1.44 & 10.13 \\
                        CLLA & 4160 & 16 & 16 & 96 & 1+1 / 1+16 & 1.44 & 10.84 \\
                        CLLA-quant & 4160 & 16 & 16 & 96 & 1+1 / 1+16 & 1.44 & 10.84 \\
                        \bottomrule
                \end{tabular}
        }
        \caption{Model size and hyper-parameters used for main experiments in Section \ref{experiments}.}
        \label{tab:model_detail}
\end{table*}

All training runs were conducted using 64 NVIDIA H800 GPUs, with inference evaluations performed on the same hardware configuration. All the training settings as specified in Table \ref{tab:train hyperparameters}.

\begin{table*}[t]
        \centering
                \begin{tabular}{lc}
                        \toprule
                        Hyper-parameters & Values\\
                        \midrule
                        Optimizer & AdamW \\
                        Adam $\beta$ & (0.9, 0.95) \\
                        Adam $\epsilon$ & $1e^{-8}$ \\
                        Weight Decay & 0.1 \\
                        LR & $4.2e^{-4}$ \\
                        Min LR & $4.2e^{-5}$ \\
                        LR Decay Style & Cosine \\
                        Clip Grad & 1.0 \\
                        Batch Size & 4M Tokens \\
                        Warmup Steps & 2,000 \\
                        MoE Loss Coefficient & 0.005 \\
                        Moe Capacity Factor & 1.5 \\
                        Mixed Precision & BF16 \\
                        \bottomrule
                \end{tabular}
        \caption{Training hyper-parameters used for main experiments in Section \ref{experiments}.}
        \label{tab:train hyperparameters}
\end{table*}

\section{Evaluation Datasets Details}
\label{app:datasets detailss}
\begin{itemize}
    \item \textbf{MMLU} \cite{hendrycks2020measuring}: A large-scale dataset for evaluating a model's understanding across multiple languages and domains. Models are evaluated on the VAL split with 5 shot.
    \item \textbf{Hellaswag} \cite{zellers2019hellaswag}: A Greek-language variant of the SWAG dataset, assessing a model's ability to reason about everyday situations.(zero shot)
    \item \textbf{ARC} \cite{clark2018think}: A dataset designed to test a model's ability to answer questions that require complex reasoning and commonsense knowledge.(zero shot)
    \item \textbf{Winogrande} \cite{sakaguchi2021winogrande}: An extension of the Winograd Schema Challenge, focusing on pronoun disambiguation in a large-scale setting.(zero shot)
    \item \textbf{BoolQ} \cite{clark2019boolq}: A dataset consisting of yes/no questions, where the answers can be inferred from a given passage.(zero shot)
    \item \textbf{PIQA} \cite{bisk2020piqa}: A dataset that combines visual and textual understanding, where models must answer questions about images based on a persona.(zero shot)
    \item \textbf{SIQA} \cite{sap2019socialiqa}: A dataset that evaluates a model's ability to understand social scenarios and make appropriate inferences.(zero shot)
    \item \textbf{CMMLU} \cite{li2023cmmlu}: A Chinese version of the MMLU dataset, focusing on multitask language understanding in Chinese. Models are evaluated on the VAL split with 5 shot.
    \item \textbf{CEval} \cite{huang2024c}: A benchmark for evaluating Chinese natural language processing tasks with 5 shot.
    \item \textbf{CLUEWSC2020} \cite{xu2020clue}: A Chinese dataset for named entity recognition in the financial domain.(zero shot)
    \item \textbf{Thuc-news} \cite{sun2016thuctc}: A dataset containing Chinese news articles, used for text classification and other NLP tasks.(zero shot)
    \item \textbf{OCNLI} \cite{hu2020ocnli}: A Chinese dataset for natural language inference tasks.(zero shot)
    \item \textbf{C3} \cite{sun2020investigating}: A dataset designed to assess a model's understanding of commonsense knowledge in Chinese.(zero shot)
    \item \textbf{FewCLUE-CHID} \cite{xu2021fewclue}: Few-shot Chinese Language Understanding Evaluation - Chinese Health Information Dataset. A dataset for few-shot learning in the healthcare domain in Chinese.(zero shot)
    \item \textbf{CMNLI}: Chinese multilingual natural language inference dataset in CLUE \cite{xu2020clue}, specifically designed for natural language inference tasks.(zero-shot)
\end{itemize}

\section{Results for Ablation Experiments}
\label{app:ablation exp detailss}
We provide detailed evaluation scores for the models discussed in Section \ref{ablation}, which include benchmarks for both English and Chinese tasks in Table \ref{tab:ablation en benchmarks} and Table \ref{tab:albation zh benchmarks}.

\begin{table*}[t]
        \centering
        \resizebox{\linewidth}{!}{
                \begin{tabular}{lccccccccc}
                        \toprule
                        Model & MMLU & Hellaswag & ARC & Winogrande & BoolQ & PIQA & SIQA & AVG\\
                        \midrule
                        CLLA+share-krope & 35.85 & 54.01 & 50.04 & 56.45 & 67.47 & 75.97 & 53.87 & 56.24 \\
                        CLLA+share-kvproj & 42.99 & 54.40 & 58.86 & 57.38 & 71.67 & 74.86 & 56.14 & 59.47 \\
                        CLLA-quant+share-kvproj & 42.15 & 54.53 & 57.86 & 56.99 & 68.23 & 74.54 & 54.76 & 58.44 \\
                        \bottomrule
                \end{tabular}
        }
        \caption{Detailed English benchmarks evaluations for ablation experiments in Section \ref{ablation}.}
        \label{tab:ablation en benchmarks}
\end{table*}

\begin{table*}[t]
        \centering
        \resizebox{\linewidth}{!}{
                \begin{tabular}{lcccccccccc}
                        \toprule
                        Model & CMMLU & CEval & CLUEWSC2020 & Thuc-news & OCNLI & C3 & CHID & CMNLI & AVG\\
                        \midrule
                        CLLA+share-krope & 29.03 & 36.08 & 71.07 & 96.92 & 57.04 & 63.21 & 81.70 & 58.62 & 61.71 \\
                        CLLA+share-kvproj & 41.50 & 44.40 & 70.19 & 96.10 & 47.39 & 60.83 & 79.70 & 51.08 & 61.40 \\
                        CLLA-quant+share-kvproj & 39.17 & 41.70 & 71.15 & 96.80 & 48.78 & 60.97 & 78.22 & 54.83 & 61.45 \\
                        \bottomrule
                \end{tabular}
        }
        \caption{Detailed Chinese benchmarks evaluations for ablation experiments in Section \ref{ablation}.}
        \label{tab:albation zh benchmarks}
\end{table*}

\end{document}